\pdfoutput=1
\documentclass[10pt,twocolumn,letterpaper]{article}

\usepackage{iccv}
\usepackage{times}
\usepackage{epsfig}
\usepackage{graphicx}
\usepackage{amsmath}
\usepackage{amssymb}

\usepackage{caption} 
\usepackage{float}
\usepackage{bm}
\usepackage{upgreek}
\usepackage[flushleft]{threeparttable}
\usepackage{array}
\newcolumntype{P}[1]{>{\centering\arraybackslash}p{#1}}
\usepackage{subcaption}
\usepackage{tabularx}
\usepackage[T1]{fontenc}
\usepackage{booktabs}
\usepackage{multirow}
\usepackage{array}
\usepackage{bbding}
\usepackage{makecell}
\usepackage{enumitem}
\usepackage{authblk}
\newcommand*\samethanks[1][\value{footnote}]{\footnotemark[#1]}

\usepackage[breaklinks=true,bookmarks=false]{hyperref}

\iccvfinalcopy 

\def\iccvPaperID{****} 

\ificcvfinal\fi

\begin{document}

\title{LDDMM-Face: Large Deformation Diffeomorphic Metric Learning for Flexible and Consistent Face Alignment}

\author[1,2]{Huilin Yang \thanks{Equal contribution}}
\author[1,3]{Junyan Lyu \samethanks}
\author[1]{Pujin Cheng}
\author[1]{Xiaoying Tang}

\affil[1]{Southern University of Science and Technology \authorcr
  \tt tangxy@sustech.edu.cn}
\affil[2]{The University of British Columbia}
\affil[3]{The University of Queensland}

\maketitle
\ificcvfinal\thispagestyle{empty}\fi

\begin{abstract}
   We innovatively propose a flexible and consistent face alignment framework, LDDMM-Face, the key contribution of which is a deformation layer that naturally embeds facial geometry in a diffeomorphic way. Instead of predicting facial landmarks via heatmap or coordinate regression, we formulate this task in a diffeomorphic registration manner and predict momenta that uniquely parameterize the deformation between initial boundary and true boundary, and then perform large deformation diffeomorphic metric mapping (LDDMM) simultaneously for curve and landmark to localize the facial landmarks. Due to the embedding of LDDMM into a deep network, LDDMM-Face can consistently annotate facial landmarks without ambiguity and flexibly handle various annotation schemes, and can even predict dense annotations from sparse ones. Our method can be easily integrated into various face alignment networks. We extensively evaluate LDDMM-Face on four benchmark datasets: 300W, WFLW, HELEN and COFW-68. LDDMM-Face is comparable or superior to state-of-the-art methods for traditional within-dataset and same-annotation settings, but truly distinguishes itself with outstanding performance when dealing with weakly-supervised learning (partial-to-full), challenging cases (e.g., occluded faces), and different training and prediction datasets. In addition, LDDMM-Face shows promising results on the most challenging task of predicting across datasets with different annotation schemes.
\end{abstract}

\begin{figure}[thbp]
\begin{center}
  \includegraphics[width=1\linewidth]{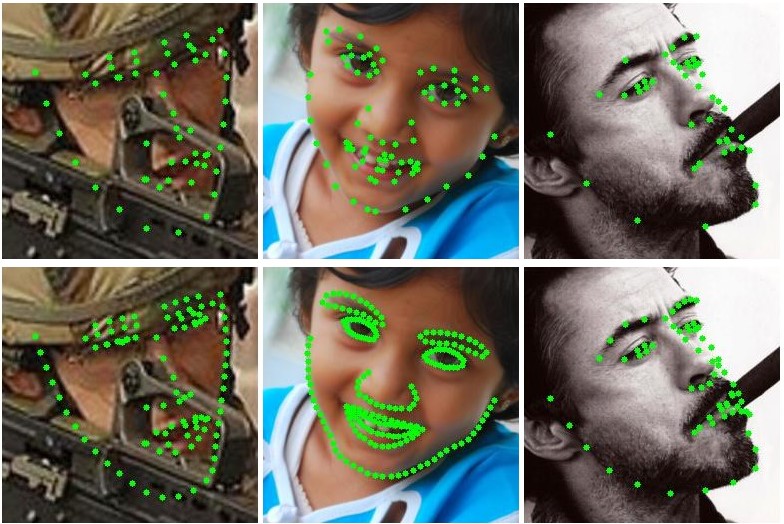}
\end{center}
  \caption{Representative LDDMM-Face results. Top: training with partial annotations on the WFLW, HELEN and 300W datasets (from left to right); bottom: predictions of full annotations from LDDMM-Face.}
\label{fig:intro}
\end{figure}

\section{Introduction}
Face alignment refers to identifying the geometric structure of a human face in a digital image, through localizing key landmarks that are usually predefined and characterizable of the face's geometry. Face alignment is a prerequisite in many computer vision tasks, such as face recognition \cite{Yi_2013_CVPR},  facial expression recognition \cite{sariyanidi2015automatic, yang2018geometry}, face verification \cite{taigman2014deepface}, face reconstruction \cite{richardson2017learning} and face reenactment \cite{nirkin2019fsgan}. 

For different datasets, there exist different face alignment annotation schemes. For example, COFW \cite{burgos2013robust} annotates 29 landmarks, 300W \cite{sagonas2013300} annotates 68 landmarks,  WFLW \cite{Wu_2018_CVPR} annotates 98 landmarks, and HELEN \cite{le2012interactive} annotates 194 landmarks. Most existing face alignment methods can only deal with the specific annotation scheme adopted by the training dataset of interest, but cannot flexibly accommodate multiple annotation schemes. Namely, if a model is trained on a dataset with a specific annotation scheme, it can then only predict landmarks of the specific scheme; a model trained on 300W with a 68-landmark annotation scheme can only predict the learned 68 landmarks but not other annotation schemes such as a 194-landmark scheme. In addition, to date, there is no such work that can fully utilize partially-annotated data. In other words, it is infeasible to make full predictions based on only partially-annotated data; for example, predicting 194 landmarks when training with only 97 or even less landmarks. Even in traditional fully-supervised settings, most existing works cannot very well handle challenging cases such as faces with occlusion. This is because those existing works usually handle each landmark individually or predict landmarks with non-diffeomorphic deformations. In this way, the predicted landmarks could be inconsistent, resulting in incorrect facial geometry topology.

In such context, we 
formulate face alignment into a diffeomorphic registration framework. Specifically, we use 
 boundary curves to represent facial geometry \cite{dupuis1998variational}. Then, large deformation diffeomorphic metric mapping (LDDMM) simultaneously for curve and landmark \cite{glaunes2008large,joshi2000landmark} between an initial face and the true face is encoded into a neural network for landmark localization. LDDMM delivers a non-linear smooth transformation with a favorable topology-preserving one-to-one mapping property. Once the diffeomorphism, which is parameterized by momenta \cite{dupuis1998variational,glaunes2008large,joshi2000landmark}, between the initial face and the true face is obtained, all points on/around the initial face have unique correspondence on/around the true face through the acquired diffeomorphism. This property makes it possible to predict facial landmarks of different annotation schemes with a model trained only on landmarks from a single annotation scheme. Utilizing both landmark and curve enables LDDMM to handle shape deformations both locally and globally; the role of the landmark term is to match the corresponding landmarks whereas the role of the curve term is to make the corresponding facial curves be close to each other and to preserve facial topology such that consistent landmark predictions can be made. Noteworthily, we predict momenta instead of increments between the initial face and the true face, which gives it the additional flexibility and is the key novelty of the proposed method. This is the first time that face alignment is formulated as a diffeomorphic registration problem, providing novel insights into the face alignment realm.

In this work, our contributions include: 
\begin{itemize}
\item[•] We propose a novel face alignment network by integrating LDDMM into deep neural networks to handle various facial annotation schemes. Our proposed approach, LDDMM-Face, can be easily integrated into most face alignment networks to effectively predict facial landmarks with different annotation schemes.
\item[•] Our approach for the first time identifies the feasibility of predicting consistent facial boundaries and full facial landmarks supervised with only partial annotations of training data, in a novel manner of weakly-supervised learning.
\item[•] We comprehensively evaluate the performance of our proposed LDDMM-Face on multiple widely-used face alignment benchmark datasets \cite{le2012interactive, Wu_2018_CVPR, sagonas2013300}, in terms of not only landmark prediction accuracy but also overall facial geometry matching degree.
\item[•] We demonstrate the effectiveness of LDDMM-Face in being superior in handling challenging cases even across datasets, being excellent in making partial-to-full predictions, being adaptable to various deep network settings, being capable of predicting consistent facial boundaries with different training annotations, and being flexible in handling multiple annotation schemes either within or across datasets. 
\end{itemize} 

\section{Related Works}\label{related works}
Face alignment has been widely researched with fruitful outcomes. From early Active Appearance Models \cite{cootes2001active,matthews2004active,kahraman2007active,saragih2007nonlinear}, Active Shape Models \cite{milborrow2008locating} to recently-developed Cascaded Shape Regression
\cite{xiong2013supervised,su2019soft,kazemi2014one,ren2014face,lee2015face,zhu2015face,chen2014joint,tuzel2016robust} and deep learning methods \cite{zhou2013extensive,zhang2016learning,trigeorgis2016mnemonic,xiao2016robust,fan2016approaching,bulat2016two,yang2017stacked,kowalski2017deep,newell2016stacked}, accuracy has been improved significantly. Equipped with powerful image feature extraction capability of Convolutional Neural Network (CNN), deep learning methods hold state-of-the-art (SOTA) results. The proposed method of this work, LDDMM-Face, fits in the deep learning scope.

{\bf Coordinate regression models}
Coordinate regression models use neural networks to directly predict coordinates of facial landmarks. Zhang \etal \cite{zhang2016learning} incorporates a variety of facial attributes like gender, expression and appearance in a multi-task learning framework. 
Kowalski \etal \cite{kowalski2017deep} presents the Deep Alignment Network (DAN) that employs landmark heatmap and transformation of face and landmarks to a canonical plane in CNN. 
Qian \etal \cite{qian2019aggregation} leverages disentangled style and shape space of each individual to augment existing structures via style translation, which can be regarded as a data augmentation approach. 

{\bf Heatmap regression models}
Heatmap regression models use neural networks to regress a set of heatmaps, one for each individual landmark, and then indirectly estimate landmarks from the heatmaps. Wu \etal \cite{Wu_2018_CVPR} uses facial boundary heatmap to supervise landmark prediction, utilizing a hourglass module \cite{newell2016stacked} and a message passing scheme \cite{chu2016structured}. Zou \etal \cite{zou2019learning} employs hierarchically structured landmark ensembles to depict holistic and local structures of facial landmarks for robust facial landmark detection. Wang \etal \cite{wang2019adaptive} proposes an adaptive wing loss for heatmap regression, with an ability to adapt its shape to various types of ground truth heatmap pixels. Kumar \etal \cite{kumar2020luvli} uses a mean estimator for heatmap and jointly predicts landmark locations, associating uncertainties of the predicted locations and landmark visibilities. Browatzki \etal \cite{browatzki20203fabrec} leverages implicit knowledge by training an autoencoder on unannotated facial datasets to perform few-shot face alignment.
 
{\bf Weakly supervised learning for face alignment} 
Weakly supervised face alignment works \cite{browatzki20203fabrec, qian2019aggregation, dong2019teacher} usually explore in a way that train models on partial datasets with full annotations. To our best knowledge, no previous works have ever made such attempts of either training on partial annotations but making full predictions or training on full annotations but predicting additional landmarks.
  
{\bf Cross-dataset/annotation face alignment}
Cross-dataset face alignment refers to training on a dataset of a specific annotation scheme while evaluating on other datasets of the same annotation scheme. Cross-annotation face alignment refers to training on a dataset of a specific annotation scheme but evaluating on datasets of different annotation schemes. Both cross-dataset and cross-annotation performance can measure a method's generalization ability. So far, existing works \cite{Wu_2018_CVPR, zhu2014transferring, zhang2015leveraging} typically utilize information from multiple different datasets and their corresponding annotation schemes to boost the training performance on one specific dataset, and no work has ever investigated flexible and consistent face alignment across datasets nor across annotations. 

LDDMM is a SOTA registration framework that has been widely used in the biomedical image field \cite{miller2015amygdalar,tang2015diffeomorphometry,jiang2018deformation,yang2017analysis}. Recently, LDDMM has also shown its effectiveness in facial recognition related fields \cite{yang2018geometry}. A key component of LDDMM is that it yields a diffeomorphism between two manifolds of interest, which inspires our proposed deformation layer in LDDMM-Face and makes it feasible to consistently predict additional landmarks (in addition to the training ones) and to make cross-annotation predictions as well as effectively deal with challenging cases.

\section{Deep LDDMM Network}\label{DeepLDDMM}
\begin{figure*}[tbhp]
\begin{center}
\includegraphics[width=6.5in]{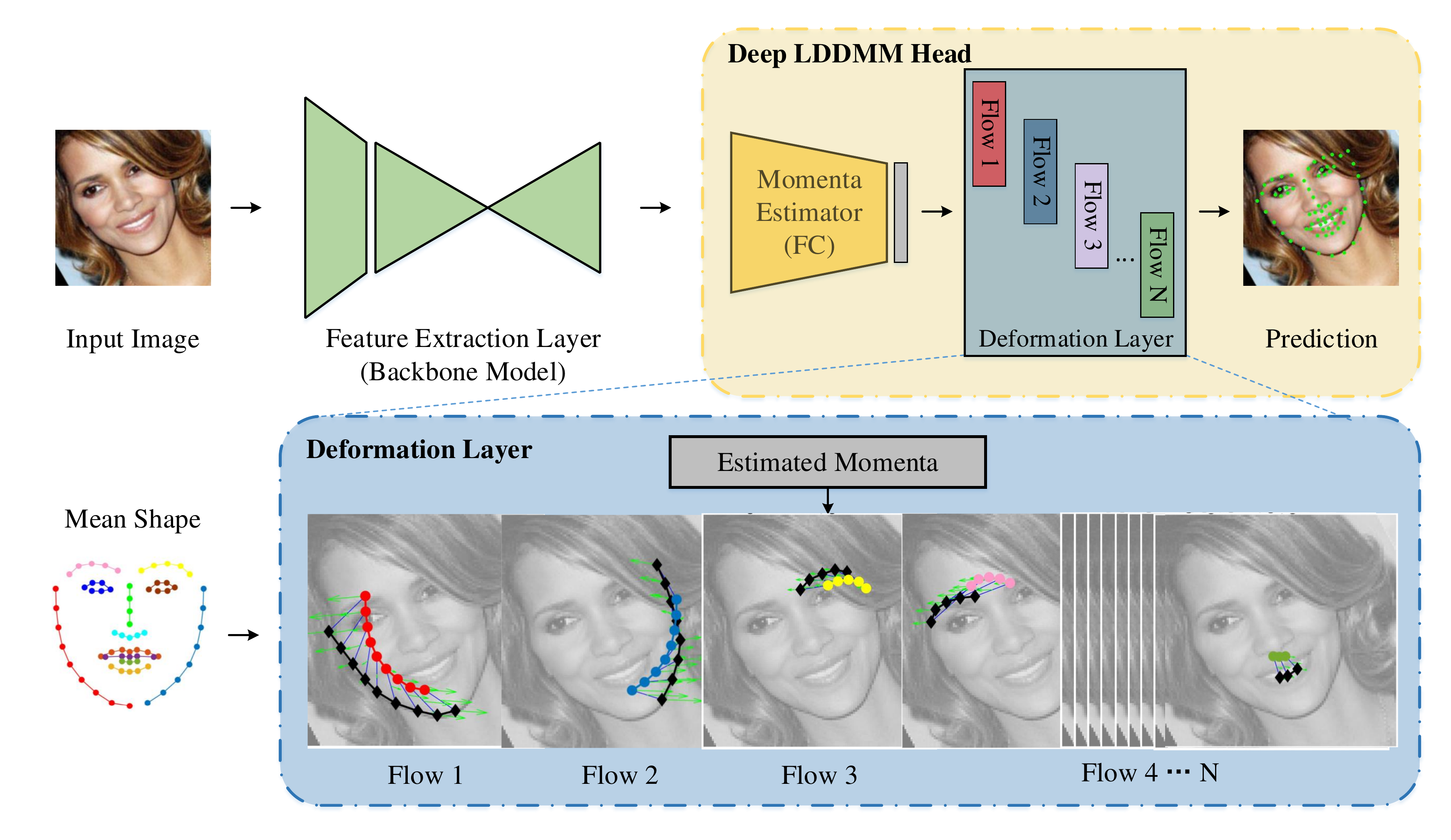}
\end{center}
\vspace*{-0mm}
   \caption{The overall pipeline of LDDMM-Face, which consists of a backbone model and two functional layers: a momenta estimator and a deformation layer that consists of N flows. In each flow of the deformation layer, the initial facial curve is shown in the same color as that in the mean face, and the deformed facial curve is shown in black connected diamonds. The fine blue lines connecting each initial landmark and the corresponding deformed landmark denote the trajectory of the initial landmarks. Green arrows show the predicted momenta at each time step along the trajectory.}
\label{fig:workflow}
\vspace*{-2mm}
\end{figure*}

As mentioned in section \ref{related works}, coordinate and heatmap regression methods cannot predict facial landmarks of different annotation schemes without retraining. In LDDMM-Face, we integrate LDDMM based facial shape registration into a deep learning framework, which can not only consistently predict facial landmarks across different annotations and datasets but also train a face alignment model in a weakly-supervised partial-to-full fashion.

Given a normalized RGB face image, LDDMM-Face first extracts both spatial and semantic features from the input image with a replaceable backbone model. Second, the features are passed through a deep LDDMM head which consists of a momenta estimator and a deformation layer. The momenta estimator contains fully-connected layers and predicts vertical and horizontal momenta for each landmark. Suppose the geometry of a face is characterized by $N$ boundary curves, the deformation layer has $N$ sublayers (flow $1$ to flow $N$). Each sublayer separately deforms the corresponding initial curve, the procedure of which is detailed in subsection~\ref{deformation layer}. Two inputs, the mean face serving as the initial face and the estimated momenta, are fed into the deformation layer. The deformed facial curves from each layer are sequentially concatenated, yielding an estimate of the true face. Fig.~\ref{fig:workflow} shows the overall pipeline of LDDMM-Face.

The structure and configurations of the baseline backbone model are identical to an existing SOTA facial landmark detector \cite{wang2020deep}. We focus on the proposed deformation layer and loss function since these components can be readily integrated into most deep learning-based face alignment pipelines and detailed investigations of the baseline network go beyond the scope of this work.

\subsection{LDDMM Deformation Layer}\label{deformation layer}

\subsubsection{LDDMM-curve$\&$landmark}

Our proposed deformation layer, based on LDDMM-curve$\&$landmark, combines the advantages of LDDMM-curve \cite{glaunes2008large} and LDDMM-landmark \cite{joshi2000landmark} to account for both global and local discrepancies in the matching process. LDDMM \cite{glaunes2008large,joshi2000landmark,dupuis1998variational} is a registration framework that provides a diffeomorphic transformation acting on the ambient space. Under the LDDMM framework, objects are placed into a reproducing kernel Hilbert metric space through time-varying velocity vector fields $v_{t}(\cdot)$: $\mathbb{R}^{2} \rightarrow \mathbb{R}^{2}$ for $t\in[0,1]$ in the ambient space. The underlying assumption is that the two objects of interest are of equivalent topology and one can be deformed from the other via a flow of diffeomorphisms. Given a pair of objects $C$ and $S$, the time-dependent flow of diffeomorphisms transforming $C$ to $S$ is defined according to the ordinary differential equation (ODE) $\dot{\phi_t}(x)=v_t (\phi_t(x))$, with $\phi_0(x)=x$. The resulting diffeomorphism $\phi_1(x)$ is acquired as the end point of the diffeomorphism flow at time $t=1$ such that $\phi_1\cdot C=S$. To ensure the resulting transformation is diffeomorphic, $v_{t}$ must satisfy the constraint that $\int_{0}^{1} {\vert\vert v_t \vert\vert_V} dt<\infty$, with $V$ being a Hilbert space associated with a reproducing kernel function $k_{V}$ and a norm $\Vert \cdot \Vert_V$ \cite{trouve1995infinite}. In practice, Gaussian kernel is selected for $k_{V}$ being
$k_{V}(a,b)=exp(-\frac{\left \| a-b \right \|^2_2}{\sigma_V^2 })$, where $\sigma_V$ represents the kernel size that is usually selected empirically and $\left \| \  \right \|_2$ denotes the $l^{2}$-norm.

In LDDMM-curve, a curve $C_{c}$ is discretized into a sequence of $n$ ordered points $\bm{\mathrm{x}}=(x_{i})_{i=1}^{n}$. That curve can be encoded by those points along with their tangent vectors such that $C_{c}={(c_{\bm{\mathrm{x}},i},\tau_{\bm{\mathrm{x}},i})}_{i=1}^{n}$, with $c_{\bm{\mathrm{x}},i}=\frac{x_{i+1}+x_{i}}{2}$ being the center of two sequential points and $\tau_{\bm{\mathrm{x}},i}=x_{i+1}-x_{i}$ being the tangent vector at point $c_{\bm{\mathrm{x}},i}$. $C_{c}$ is associated with a sum of vector-valued Diracs, $\mu_{C_{c}}=\sum_{i=1}^{n}\tau_{\bm{\mathrm{x}},i}\delta_{c_{\bm{\mathrm{x}},i}}$, and is embedded into a Hilbert metric space $W$ of smooth vectors with the norm being 

\begin{equation} \begin{split}
{\left \| \mu_{C_{c}} \right \|}_{W^{*}}^2 &={\left \| \sum_{i=1}^{n} \tau_{\bm{\mathrm{x}},i}\delta_{c_{\bm{\mathrm{x}},i}} \right \|}_{W^{*}}^2 \\
&=\sum_{i=1}^{n}\sum_{j=1}^{n}k_W(c_{\bm{\mathrm{x}},i},c_{\bm{\mathrm{x}},j})\tau_{\bm{\mathrm{x}},i}\cdot\tau_{\bm{\mathrm{x}},j},
\end{split} 
\end{equation}
where $k_W$ is the reproducing kernel in the space $W$ ($k_{W}$ is of the same form as that of $k_{V}$) and $W^{*}$ is the dual space of $W$. In LDDMM-landmark, a set of $n$ landmarks $C_{l}$ are represented by its Cartesian coordinates. Thus, a set of ordered points can be modelled as both curve and landmark. 

LDDMM-curve can handle the overall shape whereas LDDMM-landmark is more powerful in dealing with local information. Assume that the template object $C$ (the transforming object) and the target object $S$ (the object being transformed to) are respectively discretized as $\bm{\mathrm{x}}=(x_{i})_{i=1}^{n}$ and $\bm{\mathrm{y}}=(y_{i})_{i=1}^{n}$, and $\bm{\mathrm{z}}=(z_{i})_{i=1}^{n}$ is the deformed object $\phi_1\cdot C$, then the resulting diffeomorphism $\phi_1$ is obtained by minimizing the following inexact matching functional

\begin{equation}
\label{eq:J}
 \emph{J}_{\emph{c,s}} (v_{t}) = \min_{v_t:\dot{\phi_{t}} = v_{t} {(\phi_{t})},\phi_{0} = id} \gamma \int_{0}^{1} {\vert\vert v_t \vert\vert_V^2} dt + \emph{D} (\phi_1\cdot\emph{C},\emph{S}),
\end{equation}where $\int_{0}^{1} {\vert\vert v_t \vert\vert_V^2} dt$ can be interpreted as the energy consumed by the flow of diffeomorphisms, and the second term quantifies the overall discrepancy between the deformed object $\phi_1\cdot\emph{C}$ and the target object $S$. $\gamma$ is a weight in $\left [ 0, 1 \right ]$ serving as the trade-off coefficient between the consumed energy and the overall discrepancy. In LDDMM-curve$\&$landmark, the discrepancy consists of two parts
\begin{small}
\begin{equation}
\label{eq:D}
 \emph{D} (\phi_1\cdot \emph{C},\emph{S}) = \beta \emph{D}_c (\phi_1\cdot \emph{C}_c,\emph{S}_c) + \emph{D}_l (\phi_1\cdot \emph{C}_l,\emph{S}_l), 
\end{equation}
\end{small}where $\emph{D}_c$ measures the discrepancy between the deformed object and the target object when modelled as curves and $\emph{D}_l$ quantifies the corresponding discrepancy when modelled as landmarks. $\beta$ is a trade-off weight deciding the relative importance of curve and landmark. The curve discrepancy is computed as the norm of the difference between the two vector-valued curve representations in the space $W^{*}$, which is explicitly 

\begin{equation}
\emph{D}_c (\phi_1\cdot\emph{C}_c,\emph{S}_c)={\vert\vert \sum_{i=1}^{n} \tau_{\bm{\mathrm{z}},i}\delta_{c_{\bm{\mathrm{z}},i}} - \sum_{i=1}^{n} \tau_{\bm{\mathrm{y}},j}\delta_{c_{\bm{\mathrm{y}},j}} \vert\vert_{W^*}^2},
\label{eq:Dc}
\end{equation}
and the landmark discrepancy is computed as the Euclidean distance averaged across all point pairs
\begin{small}
\begin{equation}
\emph{D}_l (\phi_1\cdot \emph{C}_l,\emph{S}_l) =   \frac{1}{n} \sum_{i=1}^{n} \left \| \mathbf{z}_i - \mathbf{y}_i \right \| _2.
\label{eq:Dl}
\end{equation}
\end{small}

After minimizing $\emph{J}_{\emph{c,s}} (v_{t})$, the resulting diffeomorphism $\phi_1$ is parameterized by the velocity vector field $v_t(x)$ as $v_t(x)=\sum_{i=1}^{n} k_V(x_i(t),x)\alpha_i(t)$, where $\alpha_i(t)$ denotes the time-dependent momentum at the $i$-th landmark. A diffeomorphism is completely encoded by the initial momenta in the template space. These momenta can be obtained by solving the following sets of ODEs

\begin{small}
\begin{equation}
\label{traj}
\frac{dx_i(t)}{dt}=\sum_{j=1}^{n} k_V(x_j(t),x_i(t))\alpha_j(t), i=1,\dots,n,
\end{equation}
\end{small}where $x_i(t), t \in{[0,1]}$ denotes the trajectory of the $i$-th landmark on the template object.



\subsubsection{Deformation Layer}
The deformation layer takes the predicted momenta as inputs to perform LDDMM-induced deformation on the initial face. Trajectory of the $i$-th landmark is

\begin{small}
\begin{equation}
\label{eq:trans}
x_i(t)=x_i(0) + \int_{0}^{1} (\sum_{j=1}^{n} k_V(x_j(t),x_i(t))\alpha_j(t))dt.
\end{equation}
\end{small}The finally estimated true face (also called deformed face) is obtained at the end time point of the transformation flow. 

As illustrated in the top panel of Fig.~\ref{fig:workflow}, since a face is modelled using $N$ boundary curves, the LDDMM transformation component of the deformation layer is separately implemented for each curve from flow $1$ to flow $N$. $N$ depends on the annotation scheme. The procedure of each flow is demonstrated in the bottom panel of Fig.~\ref{fig:workflow}.

\subsection{Loss Function}\label{loss function}

The loss function in our proposed network is inspired by the objective function of LDDMM-curve$\&$landmark in Eq.~\ref{eq:J}. Focusing on accuracy, $\gamma$ is chosen to be 0 given that an accurate matching matters more than a geodesic path in face alignment. Although $\gamma$ is 0, the solution of the loss function is embedded into the $V$ space and still yields diffeomorphic transformations. Thus, the loss function, minimized with respect to the vector of LDDMM momenta $\alpha$, is

\begin{small} 
\begin{equation}
\min_{\mathbf{\alpha}} \frac{\sum_{p=1}^{N}   \beta \emph{D}_{pc} (\emph{S}^{\textrm{deform}}_{cp},\emph{S}^{*}_{cp}) + \emph{D}_{pl} (\emph{S}^{\textrm{deform}}_{lp},\emph{S}^{*}_{lp}) }{d_{ipd}},
\end{equation}
\end{small}where $S_{cp}^{*}$ is the vector-measured expression of the ground truth curve of the $p$-th facial curve, $S_{cp}^{\textrm{deform}}$ denotes the corresponding deformed curve, and $\emph{D}_{pc}$ quantifies the discrepancy between the ground truth and the deformed curve computed via Eq.~\ref{eq:Dc}. $S_{lp}^{*}$ is a vector representing the ground truth landmarks of the $p$-th facial curve, $S_{lp}^{\textrm{deform}}$ denotes the corresponding deformed landmarks and $\emph{D}_{pl}$ measures the discrepancy between the ground truth and the deformed landmarks computed via Eq.~\ref{eq:Dl}.  $d_{ipd}$ is the distance between the pupils of the ground truth face. $\beta$ is a trade-off coefficient between landmark and curve.

Therefore, our loss function takes discrepancies of both landmark and curve into consideration and consequentially is able to handle local as well as global discrepancies between the ground truth and the deformed face.

\subsection{Flexible and Consistent Face Alignment}\label{FCFA}
Once momenta are obtained from LDDMM between an initial face and a true face, the diffeomorphism between that face pair is uniquely defined \cite{glaunes2008large,joshi2000landmark}. This transformation can be used to deform not only landmarks used in the matching procedure but also any other landmarks sitting around the transforming face boundary. Due to the smooth, topology-preserving and one-to-one mapping property of the obtained diffeomorphism, we can compute the deformed location of any landmark lying on/around the face boundary in a consistent way. Any two deformed landmarks would never come across each other and any deformed boundary would never across itself, which is practically and intuitively reasonable for muscle motions of the human face. Suppose the initial locations of $m$ landmarks lying on/around the face boundary are $a(0)$, we have $a_k(t)=a_k(0) + \int_{0}^{1} (\sum_{j=1}^{n} k_V(x_j(t),a_k(t))\alpha_j(t))dt$, where $a_k(t)$ represents the location of the deformed $k$-th landmark at time $t$, $ k=1,\dots,m$. $x_j(t)$ and $\alpha_j(t)$ respectively denote the location and momentum of the $j$-th landmark that has been used in the matching procedure. $k_V$ is the reproducing kernel used in the matching procedure. The final locations are obtained at the end point of the transformation flow, namely $a_k(1)$. The transformations and acquired momenta are different for the $N$ facial curves, and thus landmarks on/around each curve are deformed separately.

\begin{figure}[thbp]
\begin{center}
   \includegraphics[width=1\linewidth]{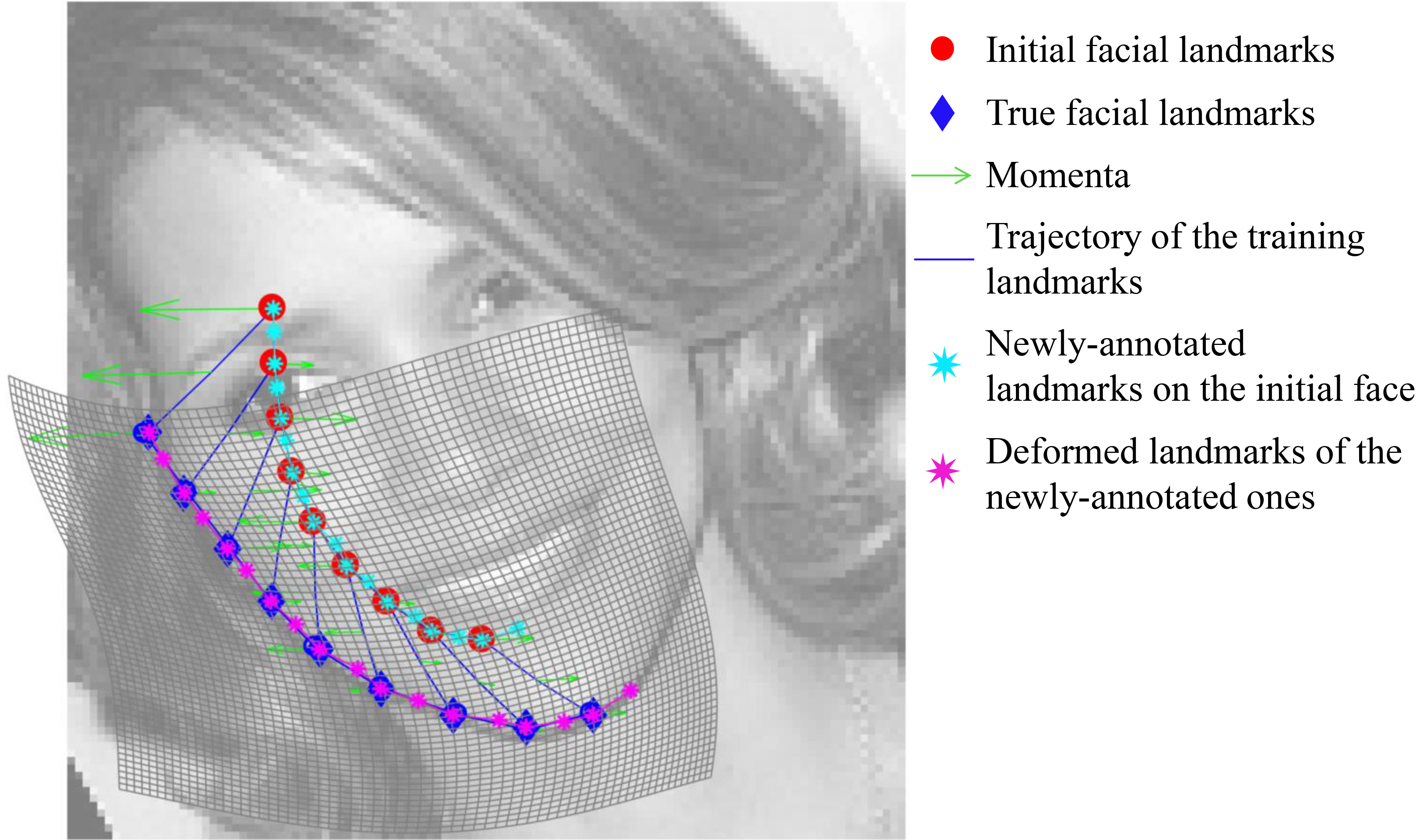}
\end{center}
   \caption{Demonstration of flexible and consistent face alignment for a right cheek. Red circles and blue diamonds respectively represent the initial and true facial landmarks involved in training. Cyan stars and magenta stars respectively represent newly-annotated landmarks on/around the initial face and the corresponding deformed landmarks of the newly-annotated ones through diffeomorphism obtained in the training stage. Green arrows denote the momenta along the trajectory of the transforming curve and blue lines represent the corresponding trajectory. Gray grids represent the diffeomorphism-induced deformations. Note that even the cyan points not used in computing the diffeomorphism are deformed correctly.}
\label{fig:LDDMM-consistent}
\vspace{-5mm}
\end{figure}

Fig.~\ref{fig:LDDMM-consistent} demonstrates an example of consistent alignment. Noticeably, some of the newly-annotated cyan star landmarks which were not involved in obtaining the LDDMM-induced diffeomorphism can still be deformed to proper locations through the predicted diffeomorphism. 

Therefore, given a pair of initial and true faces, once we have the LDDMM derived momenta, we can flexibly as well as consistently predict the deformed location of any extra landmark regardless of the training annotation scheme used. 

\section{Experiment}
In this section, the employed datasets and error metrics as well as implementation details are described below. Subsection~\ref{Adaptive} shows the adaptability of LDDMM-Face and subsection~\ref{comparison} compares our results with SOTA. Subsections~\ref{weakly-supervised} and~\ref{across-annotation} show the flexibility and consistency of LDDMM-Face by performing weakly-supervised face alignment in a partial-to-full manner, both across annotations and across datasets.

\textbf{Datasets} To evaluate the performance of LDDMM-Face, we conduct experiments on 300W \cite{sagonas2013300}, WFLW \cite{Wu_2018_CVPR}, HELEN \cite{le2012interactive} and COFW-68 \cite{burgos2013robust, ghiasi2015occlusion}, all of which are benchmark datasets for face alignment. For more details on these datasets, please refer to our supplementary material.

\textbf{Error Metrics} We use two types of metrics to quantify the face alignment error: 

\begin{itemize}
\item[-] $\textup{NME}_{\textrm{landmark}}$: The mean distance between the predicted landmarks and the ground truth landmarks divided by the inter-ocular distance \cite{ren2014face,zhu2015face,xiao2016robust}.

\item[-] $\textup{NME}_{\textrm{curve}}$: The mean iterative closest point (ICP) error between the predicted curves and the ground truth curves divided by the inter-ocular distance \cite{arun1987least}. 
\end{itemize}

Specifically, the ICP error is introduced to quantify the overall curve discrepancy and it can be used to solve the problem that inter-ocular landmark distance is unavailable when there is no point-by-point correspondence between the predicted landmarks and the ground truth landmarks. With the ICP error, fair comparisons can be conducted between a heatmap regression-based baseline method and LDDMM-Face in weakly-supervised, cross-dataset and cross-annotation face alignment settings.

Following \cite{Wu_2018_CVPR, dong2018style}, the area under the cumulative error distribution curve ($\textup{AUC}_{0.1}$) and the failure rate ($\textup{FR}_{0.1}$) are also used when evaluating the test set of WFLW. And $\textup{FR}_{0.1}$ is also used for COFW-68 \cite{wu2017leveraging, zhu2015face}. 

\textbf{Implementation} LDDMM-Face consists of a backbone model, a momenta estimator, a deformation layer and a loss function, as presented in section~\ref{DeepLDDMM}. For the backbone model, we employ three different networks for evaluating the adaptability of LDDMM-Face, which will be described in subsection~\ref{Adaptive}. For the momenta estimator, we adopt a simple yet effective structure consisting of an average pooling layer and a fully-connected layer. For the deformation layer of LDDMM-curve$\&$landmark, $\sigma_V$ and $\sigma_W$ are respectively chosen to be the scale and half the scale of the coordinates of each facial curve of the mean face. $N$ is chosen to be 12 in order to efficiently characterize different parts of a face. For the loss function, $\beta$ is empirically chosen to be 0.1. All our experiments are conducted with PyTorch 1.7.1 \cite{NEURIPS2019_bdbca288} on 4 RTX 3090 GPUs. 
More details are illustrated in our supplementary material. Codes will be released.

\subsection{Adaptive Face Alignment across Different Learning Frameworks}\label{Adaptive}
In this subsection, we investigate the adaptability and robustness of LDDMM-Face incorporated into three networks, namely HRNet \cite{wang2020deep}, Hourglass \cite{yang2017stacked} and DAN \cite{kowalski2017deep}. HRNet and Hourglass are SOTA heatmap regression methods and DAN is a multi-stage coordinate regression method. Considering computation cost, we use HRNetV2-W18, 2-stacked hourglass and VGG11 \cite{simonyan2014very} as the corresponding backbone models of the three networks. Experimental results on 300W, WFLW and HELEN (Table~\ref{Tab:1}) demonstrate that LDDMM-Face can be easily integrated into those face alignment networks. Among all three settings, LDDMM-Face (HRNet) gives the best results, which is thus employed as the default model in our subsequent experiments.




\subsection{Comparison with State-of-the-art Results}\label{comparison}
We compare LDDMM-Face with SOTA approaches on the test sets of WFLW and 300W, respectively in Table~\ref{Tab:WFLW} and Table~\ref{Tab:2}. Experimental results verify the effectiveness of LDDMM-Face. For WFLW, although this dataset confronts large variations of poses, expressions and occlusions, LDDMM-Face yields superior results and outperforms almost all compared approaches. For 300W, the performance of LDDMM-Face is comparable to its baseline HRNet and outperforms most existing methods.

\begin{table}[h!]
\renewcommand\arraystretch{1.1}
\center
\small
\begin{tabular}{@{}cccc@{}}
\toprule
\multirow{2}{*}{Method} & \multicolumn{3}{c}{$\textup{NME}_{\textrm{landmark}}$(\%)} \\ \cmidrule(l){2-4} 
 & \multicolumn{1}{c}{300W} & \multicolumn{1}{c}{WFLW} & \multicolumn{1}{c}{HELEN} \\ \midrule
\multicolumn{1}{l}{LDDMM-Face (HRNet)} & \textbf{3.53} & \textbf{4.63} & \multicolumn{1}{c}{\textbf{3.57}} \\
\multicolumn{1}{l}{LDDMM-Face (Hourglass)} & 3.73 & 5.00 & \multicolumn{1}{c}{3.89} \\
\multicolumn{1}{l}{LDDMM-Face (DAN)} & 3.91 & 5.43 & \multicolumn{1}{c}{3.95} \\ \bottomrule
\end{tabular}%
\caption{$\textup{NME}_{\textrm{landmark}}$ of LDDMM-Face incorporated into three different face alignment settings, obtained on the 300W full set, WFLW test set and HELEN test set.}
\label{Tab:1}
\vspace{-2mm}
\end{table}

\begin{table}[h!]
\renewcommand\arraystretch{1.1}
    \center
    \small
    \resizebox{\columnwidth}{!}{%
    \begin{tabular}{@{}cccc@{}}
    \toprule
    \multicolumn{1}{c}{Method} & \multicolumn{1}{c}{$\textup{NME}_{\textrm{landmark}}$(\%)} & \multicolumn{1}{c}{$\textup{FR}_{0.1}$(\%)}  & \multicolumn{1}{c}{$\textup{AUC}_{0.1}$} \\ 
    \hline
    \multicolumn{1}{l}{CFSS \cite{zhu2015face}} & 9.07 & 29.40 & \multicolumn{1}{c}{0.3659} \\ 
    \multicolumn{1}{l}{DVLN \cite{wu2017leveraging}} & 6.08 & 10.84 & \multicolumn{1}{c}{0.4551} \\
    \multicolumn{1}{l}{3FabRec \cite{browatzki20203fabrec}} & 5.62 & 8.28 & \multicolumn{1}{c}{0.4840} \\
    \multicolumn{1}{l}{LAB (Extra Data) \cite{Wu_2018_CVPR}} & 5.27 & 7.56 & \multicolumn{1}{c}{0.5323} \\
    \multicolumn{1}{l}{AVS \cite{qian2019aggregation}} & 5.25 & 7.44 & \multicolumn{1}{c}{0.5034} \\
    \multicolumn{1}{l}{SAN \cite{dong2018style}} & 5.22 & 6.32 & \multicolumn{1}{c}{0.5355} \\
    \multicolumn{1}{l}{HRNet \cite{wang2020deep}} & \textbf{4.60} & 4.64 & \multicolumn{1}{c}{0.5237} \\
    \hline
    \multicolumn{1}{l}{\textbf{LDDMM-Face}} & 4.63 & \textbf{3.68} & \multicolumn{1}{c}{\textbf{0.5509}} \\
    \multicolumn{1}{l}{\textbf{LDDMM-Face (Weak-LF: 50\%)}} & 4.79 & 4.12 & \multicolumn{1}{c}{0.5352} \\
    \bottomrule
    \end{tabular}%
    }
    \caption{$\textup{NME}_{\textrm{landmark}}$, $\textup{FR}_{0.1}$ and $\textup{AUC}_{0.1}$ results on the WFLW test set. "Weak-LF" is short for training landmark fraction in weakly-supervised face alignment.}
    \label{Tab:WFLW}
\vspace{-2mm}
\end{table}

\begin{table}[h!]
\renewcommand\arraystretch{1.1}
    \center
    \small
    \resizebox{\columnwidth}{!}{%
    \begin{tabular}{@{}cccc@{}}
    \toprule
    \multirow{3}{*}{Method} & \multicolumn{3}{c}{$\textup{NME}_{\textrm{landmark}}$(\%)} \\ \cmidrule(l){2-4} 
     & \multicolumn{1}{c}{300W} & \multicolumn{1}{c}{300W} & \multicolumn{1}{c}{300W} \\ 
     & \multicolumn{1}{c}{Common} & \multicolumn{1}{c}{Challenging} & \multicolumn{1}{c}{Full} \\ \midrule
    \multicolumn{1}{l}{MDM \cite{trigeorgis2016mnemonic}} & 4.83 & 10.14 & \multicolumn{1}{c}{5.88} \\
    \multicolumn{1}{l}{RAR \cite{xiao2016robust}} & 5.03 & 8.95 & \multicolumn{1}{c}{5.80} \\
    \multicolumn{1}{l}{SAN \cite{dong2018style}} & 3.34 & 6.60 & \multicolumn{1}{c}{3.98} \\
    \multicolumn{1}{l}{ODN \cite{zhu2019robust}} & 3.91 & 5.43 & \multicolumn{1}{c}{3.95} \\
    \multicolumn{1}{l}{3FabRec \cite{browatzki20203fabrec}} & 3.36 & 5.74 & \multicolumn{1}{c}{3.82} \\
    \multicolumn{1}{l}{DAN \cite{kowalski2017deep}} & 3.15 & 5.53 & \multicolumn{1}{c}{3.62} \\
    \multicolumn{1}{l}{LAB (Extra Data) \cite{Wu_2018_CVPR}} & 2.98 & 5.19 & \multicolumn{1}{c}{3.49} \\
    \multicolumn{1}{l}{DeCaFa (Extra Data) \cite{dapogny2019decafa}} & 2.93 & 5.26 & \multicolumn{1}{c}{3.39} \\
    \multicolumn{1}{l}{HRNet \cite{wang2020deep}} & \textbf{2.91} & \textbf{5.11} & \multicolumn{1}{c}{\textbf{3.34}} \\
    \hline
    \multicolumn{1}{l}{\textbf{LDDMM-Face}} & 3.07 & 5.40 & \multicolumn{1}{c}{3.53} \\
    \multicolumn{1}{l}{\textbf{LDDMM-Face (Weak-LF: 50\%)}} & 3.18 & 5.65 & \multicolumn{1}{c}{3.67} \\
    \bottomrule
    \end{tabular}%
    }
    \caption{$\textup{NME}_{\textrm{landmark}}$ results on the 300W common set, challenging set and full set.}
    \label{Tab:2}
\vspace{-2mm}
\end{table}

\subsection{Flexible and Consistent Face Alignment in a Weakly-supervised Manner}
\label{weakly-supervised}
In this subsection, we show the power of LDDMM-Face in two folds. We first validate the flexibility and consistency of LDDMM-Face by performing weakly-supervised learning. As described in subsection \ref{FCFA}, LDDMM-Face can predict any extra landmark lying nearby the predefined curve. Thus, we can train a model with partial landmarks that minimally describe facial geometry to predict full landmarks in a consistent way. We conduct such experiments also on 300W, WFLW and HELEN. For 300W and WFLW, 50\% facial landmarks are used for weakly-supervised training. Since HELEN has a total of 194 landmarks annotated which is a relatively large number, we further reduce the training landmarks to 33\% in the HELEN experiment. As tabulated in Table~\ref{Tab:3}, LDDMM-Face outperforms its baseline by a large margin. In respect of $\textup{NME}_{\textrm{curve}}$, there is a 40\% improvement on 300W, a 10\% improvement on WFLW and a 20\% improvement on HELEN, when trained with 50\% landmarks. A 35\% improvement is observed on HELEN when trained with 33\% landmarks. Noteworthily, LDDMM-Face is much better than the baseline in detecting face contour and eyebrow, indicating it works better for curves with large deformations. When trained on partial landmarks, there is only very mild decline in LDDMM-Face's performance compared to training on full landmarks. 

As shown in Tables~\ref{Tab:WFLW} -~\ref{Tab:2}, on both
WFLW and 300W, when trained with 50\% landmarks, weakly-supervised LDDMM-Face still holds SOTA and performs even better than some of the fully supervised methods. Other methods cannot perform such partial-to-full predictions. More results are presented in the supplementary material.

\begin{table}[h!]
\renewcommand\arraystretch{1.1}
   \center
   \small
   \resizebox{\columnwidth}{!}{%
       \begin{tabular}{c | c | c  c  c  c  c | c}  \toprule
       \multirow{2}{*}{Methods}  &
       \multicolumn{6}{c|}{$\textup{NME}_{\textrm{curve}}$ (\%)} & 
       $\textup{NME}_{\textrm{landmark}}$ (\%) \\ \cline{2-8}
       
       & O
       & F
       & E
       & N
       & I
       & M & Full landmarks (100\%) \\ \hline
    
       \multicolumn{8}{c}{300W \textit{training landmark fraction: 50\%}} \\ \hline
       \multicolumn{1}{l|}{HRNet} & 4.82 & 9.34 & 6.00 & 3.18 & \textbf{1.86} & 3.74 & - \\
       \multicolumn{1}{l|}{LDDMM-Face} & \textbf{2.94} & \textbf{4.82} & \textbf{3.47} & \textbf{2.28} & 1.87 & \textbf{2.25} & \textbf{3.18} \\ \hline
       \multicolumn{8}{c}{HELEN \textit{training landmark fraction: 50\%}} \\ \hline
       \multicolumn{1}{l|}{HRNet} & 2.95 & 4.33 & 3.08 & 3.16 & 1.77 & 2.40 & -\\
       \multicolumn{1}{l|}{LDDMM-Face}
       & \textbf{2.39} & \textbf{3.37} & \textbf{2.52} & \textbf{2.61} & \textbf{1.46} & \textbf{1.97} & \textbf{3.71} \\ \hline
       \multicolumn{8}{c}{HELEN \textit{training landmark fraction: 33\%}} \\ \hline
       \multicolumn{1}{l|}{HRNet} & 3.73 & 5.63 & 3.95 & 3.92 & 2.15 & 3.01 & - \\
       \multicolumn{1}{l|}{LDDMM-Face}
       & \textbf{2.45} & \textbf{3.29} & \textbf{2.74} & \textbf{2.76} & \textbf{1.56} & \textbf{1.91} & \textbf{3.78}\\ \hline
       \multicolumn{8}{c}{WFLW \textit{training landmark fraction: 50\%}} \\ \hline
       \multicolumn{1}{l|}{HRNet} & 3.95 & 5.72 & 4.04 & 3.34 & 3.30 & \textbf{3.36} & -\\
       \multicolumn{1}{l|}{LDDMM-Face}
       & \textbf{3.58} & \textbf{4.67} & \textbf{3.72} & \textbf{3.20} & \textbf{2.95} & 3.38 & \textbf{4.79} \\ 
       \bottomrule
       \end{tabular}
   }
   \caption{Partial-to-full prediction results on the 300W common set, HELEN test set and WFLW test set for weakly-supervised face alignment. \textit{Landmark fraction} means the fraction of full landmarks used in training stage of  weakly-supervised face alignment. 'O' indicates overall face. 'F' means face contour. 'E' means eyebrows. 'N' means nose. 'I' means eyes. 'M' means mouth. '-' indicates $\textup{NME}_{\textrm{landmark}}$ is unavailable for HRNet since it cannot make cross-annotation predictions.}
   \label{Tab:3}
   
\vspace{-2mm}
\end{table}

\subsection{Flexible and Consistent Face Alignment across Annotations and Datasets}
\label{across-annotation}
We further validate the flexibility and consistency of LDDMM-Face by evaluating cross-dataset/annotation face alignment performance. Existing cross-dataset evaluations mainly utilize the COFW-68 dataset which have been reannotated with an identical scheme as that of 300W.

\begin{figure}[thbp]
\begin{center}
  \includegraphics[width=1\linewidth]{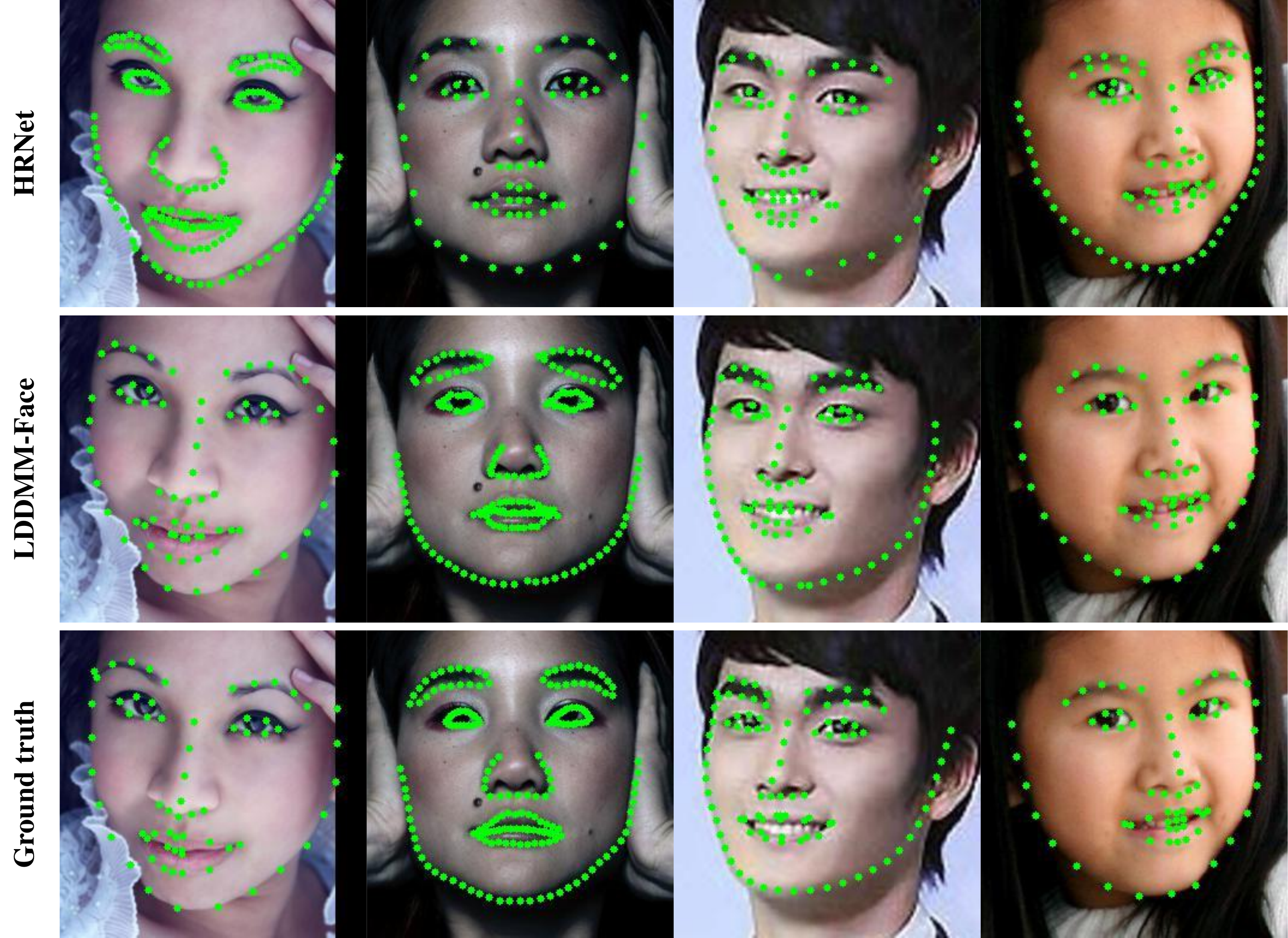}
\end{center}
  \caption{Representative cross-annotation/dataset face alignment results. From left to right, training/testing is respectively conducted on the HELEN/300W, 300W/HELEN, 300W/WFLW and WFLW/300W.}
\label{fig:cro}
\vspace{-4mm}
\end{figure}

As mentioned above, HELEN has two annotation schemes since it is also a subset of 300W. As such, the cross-annotation face alignment experiments between HELEN and 300W can be treated as cross-annotation but within-dataset. By conducting an affine transformation from source mean face to target mean face, we can easily predict landmarks of different annotation schemes without retraining. From Fig.~\ref{fig:cro} and Table~\ref{Tab:4}, we observe that LDDMM-Face significantly improves the performance over the baseline. It should be noted that although the 194-landmark annotation scheme of HELEN describes the nose and eyebrow in totally different ways from the 68-landmark annotation scheme of 300W, LDDMM-Face achieves decent performance. We also conduct simultaneous cross-dataset and cross-annotation experiments between 300W and WFLW, on which only slight improvements are observed due to the highly similar annotation schemes between these two datasets. Table~\ref{Tab:4} shows that LDDMM-Face is much better than the baseline in $\textup{NME}_{\textrm{curve}}$, but the absolute value of $\textup{NME}_{\textrm{landmark}}$ is still relatively unsatisfactory compared to traditional within-dataset and within-annotation predictions. A plausible reason is that we use an affine transformation between the two different mean faces rather than directly modify the mean face use in the specific training process, and the two mean faces may be highly inconsistent with each other. With that being said, this is to the best of our knowledge the first attempt of simultaneous cross-dataset and cross-annotation face alignment, with satisfactory performance in identifying the overall facial geometry (curve error). This observation further verifies the effectiveness and importance of LDDMM-Face.

To compare with existing SOTA cross-dataset face alignment results, we further conduct experiments on COFW-68, as summarized in Table~\ref{Tab:COFW}. LDDMM-Face significantly outperforms those compared methods, especially in terms of $\textup{FR}_{0.1}$ which is very sensitive to challenging cases like large pose and occlusion. The outstanding performance of LDDMM-Face for challenging cases is mainly due to the curve and landmark induced diffeomorphism; diffeomorphic transforming ensures the deformed facial geometry is consistent with that of the initial face such that the occluded parts can still be accurately predicted. Collectively, LDDMM-Face makes precise facial geometry predictions across different annotations (both within and across datasets), performs outstandingly for cross-dataset settings, and also effectively handles challenging cases such as occluded faces. More cross-dataset/annotation results from LDDMM-Face can be found in our supplementary material.

\begin{table}[h!]
\renewcommand\arraystretch{1.1}
   \center
   \small
   \resizebox{\columnwidth}{!}{%
   \begin{tabular}{c | c | c  c  c  c  c | c}  \toprule
   \multirow{2}{*}{Methods}  &
   \multicolumn{6}{c|}{$\textup{NME}_{\textrm{curve}}$ (\%)} & 
   $\textup{NME}_{\textrm{landmark}}$ (\%) \\ \cline{2-8}
   
   & O
   & F
   & E
   & N
   & I
   & M & O \\ \hline

   \multicolumn{8}{c}{\textit{HELEN to 300W}} \\ \hline
   \multicolumn{1}{l|}{HRNet} & 5.49 & \textbf{6.72} & 5.82 & 9.00 & 2.72 & 3.21 & - \\
   \multicolumn{1}{l|}{LDDMM-Face}
 & \textbf{4.76} & 6.81 & \textbf{4.25} & \textbf{7.45} & \textbf{2.21} & \textbf{3.09} & \textbf{5.96} \\ \hline
   \multicolumn{8}{c}{\textit{300W to HELEN}} \\ \hline
   \multicolumn{1}{l|}{HRNet} & 5.60 & 6.61 & 6.18 & 9.07 & 2.79 & 3.36 & -\\
   \multicolumn{1}{l|}{LDDMM-Face}
   & \textbf{4.13} & \textbf{3.47} & \textbf{4.79} & \textbf{7.19} & \textbf{2.41} & \textbf{2.81} & \textbf{7.58} \\ \hline
   \multicolumn{8}{c}{\textit{WFLW to 300W}} \\ \hline
   \multicolumn{1}{l|}{HRNet} & 3.91 & 5.29 & 4.62 & \textbf{3.01} & 3.14 & \textbf{3.46} & - \\
   \multicolumn{1}{l|}{LDDMM-Face}
 & \textbf{3.88} & \textbf{5.27} & \textbf{4.08} & 3.32 & \textbf{2.94} & 3.76 & \textbf{4.53} \\ \hline
   \multicolumn{8}{c}{\textit{300W to WFLW}} \\ \hline
   \multicolumn{1}{l|}{HRNet} & 6.61 & 8.65 & 6.74 & \textbf{5.63} & 6.02 & 6.02 & -\\
   \multicolumn{1}{l|}{LDDMM-Face}
  & \textbf{6.04} & \textbf{6.82} & \textbf{6.41} & 5.77 & \textbf{5.33} & \textbf{5.88} & \textbf{9.58}\\  
   \bottomrule
   \end{tabular}
   }
   \caption{Comparison between LDDMM-Face and HRNet on the 300W common set, HELEN test set and WFLW test set for cross-dataset/annotation face alignment. \textit{HELEN to 300W} means training on the HELEN train set and testing on the 300W common set. Same logic applies for others.}
   \label{Tab:4}
\vspace{-1mm}
\end{table}
   
\begin{table}[h!]
\renewcommand\arraystretch{1.1}
    \center
    \small
    \begin{tabular}{@{}ccc@{}}
    \toprule
    \multicolumn{1}{c}{Method} & \multicolumn{1}{c}{$\textup{NME}_{\textrm{landmark}}$(\%)} & \multicolumn{1}{c}{$\textup{FR}_{0.1}$(\%)} \\ 
    \hline
    \multicolumn{1}{l}{PCPR \cite{burgos2013robust}} & 8.76 & \multicolumn{1}{c}{20.12} \\
    \multicolumn{1}{l}{TCDCN \cite{zhang2016learning}} & 7.66 & \multicolumn{1}{c}{16.17} \\
    \multicolumn{1}{l}{HPM \cite{ghiasi2014occlusion}} & 6.72 & \multicolumn{1}{c}{6.71} \\ 
    \multicolumn{1}{l}{SAPM \cite{ghiasi2015using}} & 6.64 & \multicolumn{1}{c}{5.72} \\
    \multicolumn{1}{l}{CFSS \cite{zhu2015face}} & 6.28 & \multicolumn{1}{c}{9.07} \\
    \multicolumn{1}{l}{HRNet \cite{wang2020deep}} & 4.97 & \multicolumn{1}{c}{3.16} \\
    \multicolumn{1}{l}{LAB (Extra Data) \cite{Wu_2018_CVPR}} & 4.62 & \multicolumn{1}{c}{2.17} \\
    \hline
    \multicolumn{1}{l}{\textbf{LDDMM-Face}} & \textbf{4.54} & \multicolumn{1}{c}{\textbf{1.18}} \\
    \bottomrule
    \end{tabular}%
    \caption{$\textup{NME}_{\textrm{landmark}}$ and $\textup{FR}_{0.1}$ results of training on 300W and testing on the COFW-68 test set.}
    \label{Tab:COFW}
\vspace{-1mm}
\end{table}

   


\section{Conclusion}
In this work, we present and validate a novel face alignment pipeline, LDDMM-Face, that is able to perform flexible and consistent face alignment and also effectively deal with challenging cases. The flexibility and consistency delivered by LDDMM-Face arise naturally from an embedding of LDDMM into deep learning. It bridges the gap between different annotation schemes and makes the task of face alignment more flexible than existing methods which can only predict landmarks involved in annotations of the training data.
Most importantly, LDDMM-Face has a great generalization ability and can be integrated into various deep learning based face alignment networks.

{\small
\bibliographystyle{ieee_fullname}
\bibliography{egbib}
}

\end{document}